\pdfoutput=1

\documentclass[11pt]{article}

\usepackage[preprint]{acl}

\usepackage{times}
\usepackage{latexsym}

\usepackage[T1]{fontenc}

\usepackage[utf8]{inputenc}

\usepackage{microtype}

\usepackage{inconsolata}

\usepackage{graphicx}

\usepackage{color}
\usepackage{colortbl}
\definecolor{tableBlue}{rgb}{0.424,0.557,0.749}
\definecolor{lightblue}{HTML}{dfebf7}
\definecolor{persiangreen}{rgb}{0.0, 0.65, 0.58}
\definecolor{cadmiumred}{rgb}{0.89, 0.0, 0.13}
\newcommand{\tableLineColor}{\rowcolor{lightblue}}
\definecolor{increaseGreen}{rgb}{0.51,0.702,0.4}
\definecolor{decreaseRed}{rgb}{0.647,0.337,0.318}
\newcommand{\increasingData}[1]{\textcolor{persiangreen}{$\uparrow$#1}}
\newcommand{\decreasingData}[1]{\textcolor{cadmiumred}{$\downarrow$#1}}

\usepackage{amsmath}
\usepackage{amssymb}
\usepackage{mathtools}
\usepackage{amsthm}
\usepackage{pifont}
\usepackage{multirow}
\usepackage{booktabs}
\usepackage{xspace}
\usepackage{balance}
\usepackage{algorithm}
\newcommand{\methodname}{VRoPE\xspace}
\newcommand{\imageunit}{\mathrm{i}}

\title{VRoPE: Rotary Position Embedding for Video Large Language Models}

\author{
 \textbf{Zikang Liu\textsuperscript{1,2}}\thanks{Equal contribution.},
 \textbf{Longteng Guo\textsuperscript{1}}\footnotemark[1],
 \textbf{Yepeng Tang\textsuperscript{3}}\footnotemark[1],
 \textbf{Tongtian Yue\textsuperscript{1,2}}
 \\
 \textbf{Junxian Cai\textsuperscript{4}},
 \textbf{Kai Ma\textsuperscript{4}},
 \textbf{Qingbin Liu\textsuperscript{4}},
 \textbf{Xi Chen\textsuperscript{4}},
 \textbf{Jing Liu\textsuperscript{1,2}}\thanks{Corresponding author.},
\\
 \textsuperscript{1}Institute of Automation, Chinese Academy of Sciences,
 \\
 \textsuperscript{2}School of Artificial Intelligence, University of Chinese Academy of Sciences,
 \\
 \textsuperscript{3}School of Computer Science and Technology, Beijing Jiaotong University,
 \\
 \textsuperscript{4}Basic Algorithm Center, Tencent
\\
\texttt{
\{liuzikang2023,yuetongtian2022\}@ia.ac.cn, yepengtang@bjtu.edu.cn
}
\\
\texttt{
\{jasoncjxcai,kylekma,qingbinliu,jasonxchen\}@tencent.com
}
\\
\texttt{
\{longteng.guo,jliu\}@nlpr.ia.ac.cn
}
}

\begin{document}
\maketitle
\begin{abstract}
Rotary Position Embedding (RoPE) has shown strong performance in text-based Large Language Models (LLMs), but extending it to video remains a challenge due to the intricate spatiotemporal structure of video frames. Existing adaptations, such as RoPE-3D, attempt to encode spatial and temporal dimensions separately but suffer from two major limitations: positional bias in attention distribution and disruptions in video-text transitions. To overcome these issues, we propose Video Rotary Position Embedding (\methodname), a novel positional encoding method tailored for Video-LLMs. Specifically, we introduce a more balanced encoding strategy that mitigates attention biases, ensuring a more uniform distribution of spatial focus. Additionally, our approach restructures positional indices to ensure a smooth transition between video and text tokens. Extensive experiments on different models demonstrate that \methodname consistently outperforms previous RoPE variants, achieving significant improvements in video understanding, temporal reasoning, and retrieval tasks. Code is available at \url{https://github.com/johncaged/VRoPE}.
\end{abstract}

\section{Introduction}
\label{sec:intro}
In recent years, Large Language Models (LLMs) have achieved remarkable progress \cite{touvron2023llama, bai2023qwen}. 
Building on the success of LLMs, Video Large Language Models (Video-LLMs) \cite{maaz2023video, li2024llama, jin2024chat} have emerged as a powerful paradigm for video-language  understanding. These models typically integrate LLMs with pre-trained vision encoders, enabling the joint modeling of video and textual information. However, a fundamental challenge in Video-LLMs lies in effectively modeling positional relationships within video sequences. 

\begin{figure}[t]
\centering
\includegraphics[width=\linewidth]{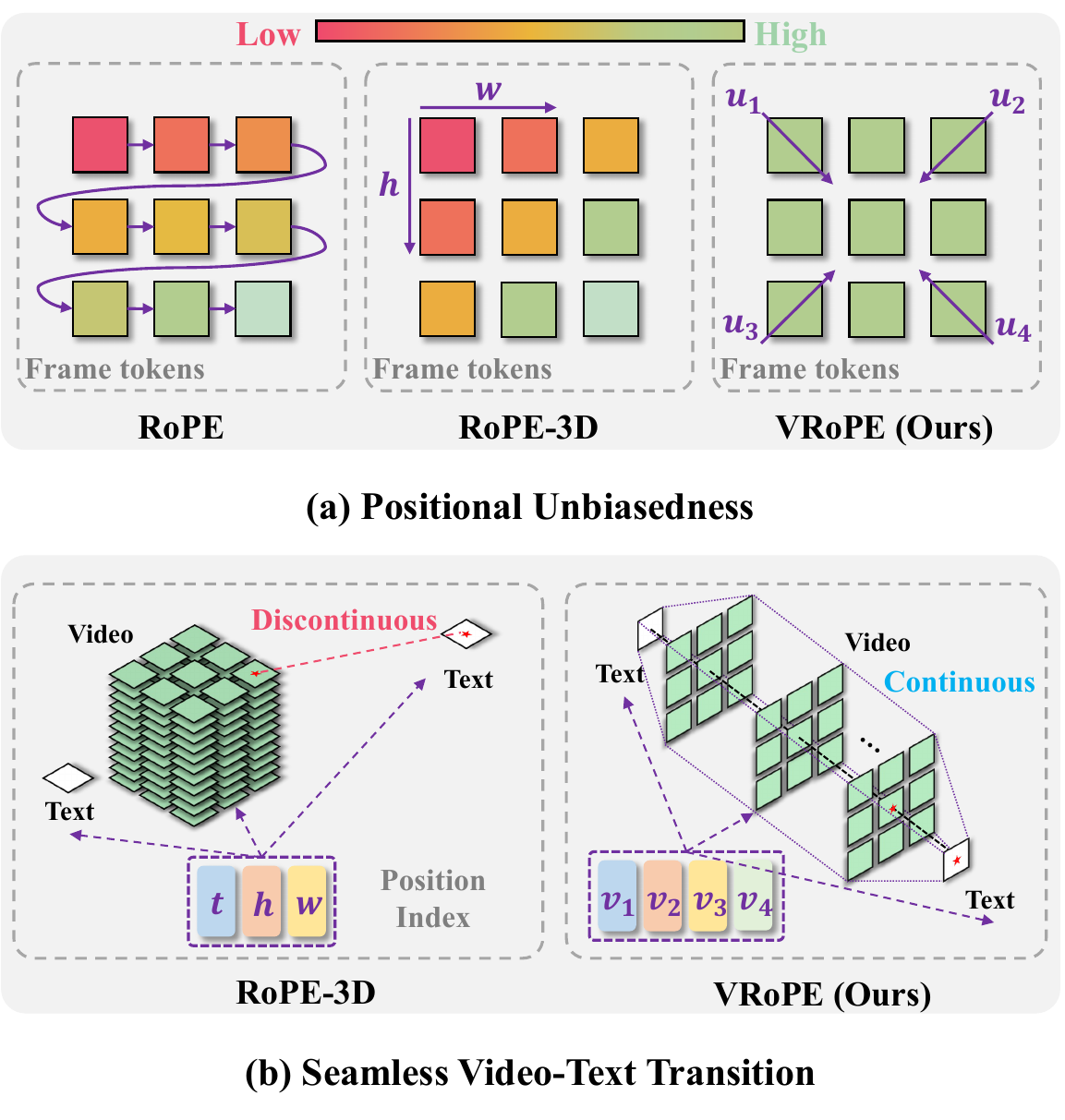}
\caption{Comparison of RoPE, RoPE-3D, and our \methodname in video positional encoding. (a) Positional Unbiasedness: RoPE and RoPE-3D exhibit spatial biased attention, particularly towards later tokens or specific frame regions, while \methodname ensures more uniform attention. (b) Seamless Video-Text Transition: RoPE-3D causes a discontinuity when transitioning from video to text tokens, which \methodname smooths for better cross-modal dependency modeling.}
\label{fig:intro}
\vspace{-2pt}
\end{figure}

In LLMs, positional encoding plays a crucial role in enabling models to capture order-dependent patterns, as self-attention mechanisms themselves are inherently permutation-invariant. Among various positional encoding schemes, Rotary Position Embedding (RoPE) \cite{su2024roformer} has gained widespread adoption due to its ability to encode relative position relationships. RoPE enables efficient long-range dependencies, making it highly effective in text-based models. However, when applied directly to video data, vanilla RoPE—where video tokens are treated as a simple sequence akin to text—fails to account for the complex spatiotemporal structure inherent in video frames, leading to suboptimal representations. Despite its critical role, an effective video-specific positional encoding strategy remains an open challenge.

To optimally encode positional relationships in Video-LLMs, we identify three key properties that an ideal video positional encoding should satisfy:

\textbf{(1) Spatiotemporal Structure Modeling.}
Unlike text, where positional relationships are strictly one-dimensional, video frames exhibit both spatial (width, height) and temporal (frame index) dimensions. An effective encoding must reflect this inherent structure to facilitate accurate modeling of spatiotemporal dependencies. 
Recent approaches \cite{wang2024qwen2, bai2025qwen2}, referred to as RoPE-3D, extend RoPE for video structure by splitting the feature channels into three parts to separately encode frame, width, and height positions.

\textbf{(2) Positional Unbiasedness.} A critical yet often overlooked aspect of positional encoding is its impact on attention distribution. As illustrated in Figure \ref{fig:intro} (a), RoPE, by design, applies a long-term decay over increasing positional indices, inadvertently introducing a bias that amplifies attention toward later tokens. This issue persists in RoPE-3D, where spatial positions within video frames are unevenly weighted, causing attention to be disproportionately focused on certain areas—typically the bottom-right regions of frames—while suppressing others, which is shown in Figure \ref{fig:intro} (a). Such biases distort spatial contextual modeling, leading to suboptimal video comprehension. An effective video positional encoding should mitigate these biases to ensure uniform attention across the entire frame.

\textbf{(3) Seamless Video-Text Transition.} For effective multimodal understanding, an ideal positional encoding should ensure a seamless transition between video and text tokens. However, as demonstrated in Figure \ref{fig:intro} (b), RoPE-3D introduces a discontinuity when transitioning from video to text tokens, as the positional indices of text tokens are arbitrarily offset by the maximum position index of the video sequence (determined by the largest of frame count, width, and height, which often vary significantly). This artificial “jump” in the positional encoding space disrupts the smooth flow of information between modalities, hindering the model to establish meaningful cross-modal dependencies.

Based on the above principles, we propose Video Rotary Position Embedding (\methodname), a novel positional encoding method specifically designed for Video-LLMs. 
Our approach consists of two key components to satisfy those principles. \textit{(1) Symmetric Bias Mitigation}: To counteract the attention bias present in RoPE-based encodings, we design a symmetric positional representation that encodes each spatial coordinate from vertices to the center. By distributing attention more uniformly across spatial locations, this method prevents positional distortions and improves overall video understanding. \textit{(2) Temporal Centered Arrangement}: We propose a center-aligned design that spatially aligns the geometric centers of video frames with the textual arrangement axis, and arranges video frames in temporally ordered progression along the textual positional axis. This transformation not only maintains spatial coherence within video frames but also ensures a smooth transition between video and text tokens, mitigating discontinuities in the positional encoding space.

Overall, \methodname effectively enhances Video-LLMs by preserving spatiotemporal structure, mitigating attention bias, and ensuring smooth video-text transitions. We conduct extensive experiments on different models and training datasets. Our results demonstrate significant performance improvements over RoPE and RoPE-3D on multiple video benchmarks, covering general video understanding, temporal reasoning, long video comprehension, and video retrieval. These findings establish \methodname as a robust and efficient positional encoding method tailored for Video-LLMs. We hope this work inspires further research on Video-LLM positional encoding and provides valuable insights for future Video-LLM designs.

\section{Related Work}

\subsection{Video Large Language Models}
Recent advancements in Video-LLMs \cite{maaz2023video, li2023videochat, li2024mvbench, jin2024chat, li2024llama, xu2024pllava} have significantly enhanced video processing by integrating multiple modalities and employing instruction fine-tuning. Notable innovations include Video-ChatGPT \cite{maaz2023video}, which introduced video instruction tuning for text generation, and VideoChat \cite{li2023videochat} and VideoChat2 \cite{li2024mvbench}, which improved modality alignment via cross-attention and multi-stage bootstrapping etc. Other models, such as Chat-UniVi \cite{jin2024chat} and LLaMA-VID \cite{li2024llama}, focus on efficient video representations through techniques like token compression and dual-token methods that separate context and content. Additionally, PLLaVA \cite{xu2024pllava} explores the use of image-pretrained LLaVA models for video tasks, utilizing simple spatial pooling techniques.

\subsection{Multimodal Position Embedding}
Most Video-LLMs inherit the default design from LLMs by using Rotary Position Embedding (RoPE) \cite{su2024roformer} for positional encoding. RoPE encodes relative distance information as absolute position embeddings, offering key advantages like no additional training parameters and improved performance in various tasks \cite{su2024roformer}. It is widely used in modern LLMs due to its ability to extrapolate context length, extending a model's window size without the need for expensive retraining. However, RoPE's 1D design, effective for text, overlooks the spatiotemporal structure of video data, limiting its suitability for Video-LLMs. To address this, several approaches have adapted RoPE for video. For instance, RoPE-2D \cite{agrawal2024pixtral, wang2024qwen2} extends the encoding to capture spatial relationships in video frames, while RoPE-3D \cite{wang2024qwen2, bai2025qwen2} divides the channel dimension into three groups to better represent the spatiotemporal dimensions.

However, these approaches still face issues like Positional Attention Bias and Cross-Modal Positional Discontinuity, which are discussed in Section \ref{sec:motivation}. Our \methodname method addresses these limitations, offering more accurate and robust positional encoding tailored for Video-LLMs.

\section{Motivation}
\label{sec:motivation}

\subsection{Preliminary: Rotary Position Embedding}
Rotary Positional Embedding (RoPE) is a widely adopted method in LLMs that encodes absolute positional information while preserving relative positional relationships. This property makes RoPE particularly effective for self-attention mechanisms, as it allows models to capture the relative distance between tokens in a computationally efficient manner.  
Given a token embedding $\mathbf{x}$ at position index $m$, RoPE applies a complex-valued rotation operation, formulated as:  
\begin{equation}
\text{RoPE}(\mathbf{x}, m) = \mathbf{x} e^{\imageunit m \boldsymbol{\theta}}
\end{equation}
\noindent where $\imageunit$ is the imaginary unit, and the frequency encoding vector $\boldsymbol{\theta}$ is defined as:
\begin{equation}
\boldsymbol{\theta}_j = \text{base}^{\frac{-2j}{d}}
\end{equation}
\noindent where $\text{base}$ is a hyperparameter, $d$ is the feature dimension, and $j = [0, 1,..., d/2 - 1] $ denotes the index of each feature channel.  

\begin{figure*}[t]
\centering
\includegraphics[width=0.9\linewidth]{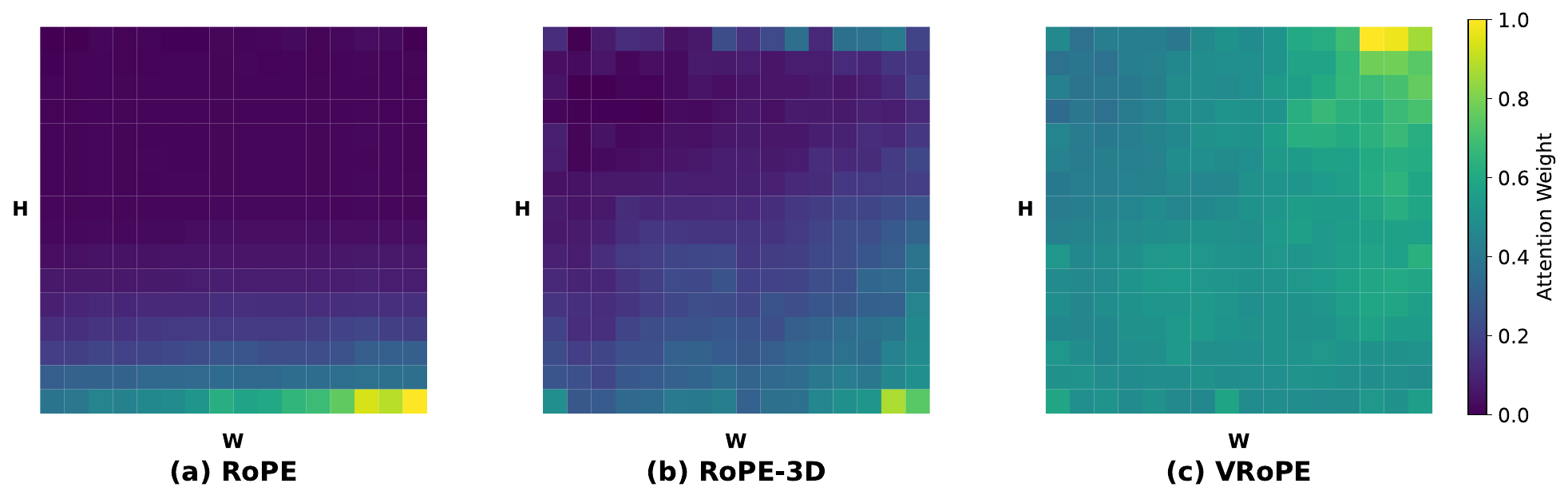}
\caption{Attention weight visualization of RoPE, RoPE-3D, and \methodname. We compute average text-to-video frame attention weights on VideoMME \cite{fu2024video} benchmark (lighter color indicates higher attention). (a) RoPE exhibits row-wise attention decay within frames. (b) RoPE-3D shows a similar decay from the bottom-right to the top-left, introducing positional bias that skews attention toward spatially closer frame tokens. (c) \methodname mitigates this bias, leading to a more balanced attention distribution.}
\label{fig:attn}
\end{figure*}

In the self-attention mechanism, RoPE transforms absolute position embeddings into relative ones.
The attention score between $m$-th query $\mathbf{q}_m$ and $n$-th key $\mathbf{k}_n$ is  
\begin{equation}
\mathbf{A}_{(m, n)} = \Re \left[ \mathbf{q}_m \cdot \mathbf{k}_n^* e^{\imageunit (m - n) \boldsymbol{\theta}} \right]
\end{equation}
\noindent where \( \Re[\cdot] \) denotes the real part, and \( ^{*} \) represents the complex conjugate.

While RoPE excels in sequential text modeling, its direct application to video-text interleaved sequences poses challenges due to the complex spatiotemporal relationships inherent in video frames.

\subsection{RoPE for Video-LLMs}

In Video-LLMs, video frames are typically processed by vision encoders (e.g., ViTs \cite{alexey2020image} or CNNs \cite{he2016deep}) and transformed into a sequence of visual tokens. These visual tokens are then concatenated with text tokens and fed into an LLM backbone.

In most existing approaches, video tokens are treated as a simple 1D sequence, with position indices assigned in an increasing order, similar to text. 
However, this naive approach, referred to as RoPE, overlooks the inherent spatiotemporal structure of video data. 
Flattening video frames this way disrupts spatiotemporal structure and leads to inefficient position usage. Unlike text, video tokens carry less dense semantic information, and their excessive sequence length can weaken contextual dependencies, making long-range understanding harder.

\subsection{RoPE-3D for Video-LLMs}

Recent approaches, such as M-RoPE in Qwen2-VL\cite{wang2024qwen2}, have proposed RoPE-3D as an extension of RoPE for video structure preserving. RoPE-3D intuitively partitions the feature dimensions to separately encode spatial (width, height) and temporal (frame index) positions.
Given a video token with coordinates $ (w, h, t) $, RoPE-3D computes:

\begin{small}
\begin{equation}
\label{eq:rope3d}
\begin{aligned}
& \text{RoPE-3D}_{j}(\mathbf{x}, w, h, t) = \left\{
\begin{aligned}
& \text{RoPE}_{j}(\mathbf{x}, w ), j \in D_{w} \\
& \text{RoPE}_{j}(\mathbf{x}, h ), j \in D_{h} \\
& \text{RoPE}_{j}(\mathbf{x}, t ), j \in D_{t}
\end{aligned}
\right.
\end{aligned}
\end{equation}
\end{small}

\noindent where where $ D_w, D_h, D_t$ denote feature partitions assigned to width, height, and temporal axes, respectively. 
For text tokens, the encoding remains consistent with the original RoPE by setting $w = h = t = m$, ensuring that:  
\begin{equation}
\label{eq:text_inv}
\text{RoPE-3D}_{j}(\mathbf{x}, m, m, m) \equiv \text{RoPE}_{j}(\mathbf{x}, m)
\end{equation}

This design explicitly models spatial and temporal positions while preserving text token behavior. However, RoPE-3D still exhibits two key limitations, which we elaborate on below.

\subsection{Problem Analysis}
\label{sec:problem}
While RoPE-3D introduces a promising design by partitioning the feature dimensions to encode spatial (width, height) and temporal (frame index) positions separately, two critical issues persist when handling video–text data.

\begin{table}[tbp]
\centering
\caption{Average attention weights at the video-text boundary on Video-MME. We use the subsequent text instruction as the query and video/text tokens as keys. Note that text-to-video attention weights of RoPE-3D are an order of magnitude lower than other methods, indicating its positional discontinuity between video and text.}
\label{tab:discontinuity}
\resizebox{0.95\linewidth}{!}{
\begin{tabular}{lcc}
\toprule
Method & Text-to-Text & Text-to-Video \\
\midrule
RoPE & 1.41e-2 & 2.08e-4 \\
RoPE-3D & 1.27e-2 & \textbf{\textcolor{cadmiumred}{5.12e-5}} \\
\methodname (Ours) & 1.32e-2 & 3.70e-4\\
\bottomrule
\end{tabular}
}
\end{table}

\paragraph{(1) Positional Attention Bias.}
As is demonstrated in Figure \ref{fig:attn} (a), RoPE naturally applies a long-term decay over increasing positional indices, which amplify attention toward later positions. Unfortunately, we find that this issue persists in RoPE-3D, where the decay leads to \textit{an uneven distribution of focus across spatial positions in video frames}. As is shown in Figure \ref{fig:attn} (b), notably, tokens in the bottom-right of each frame receive disproportionately higher attention, while those in the top-left are increasingly suppressed. This imbalance can distort spatial contextual modeling by weakening dependencies on earlier tokens, which in turn affects the model's understanding of the video.

\begin{figure*}[t]
\centering
\includegraphics[width=\linewidth]{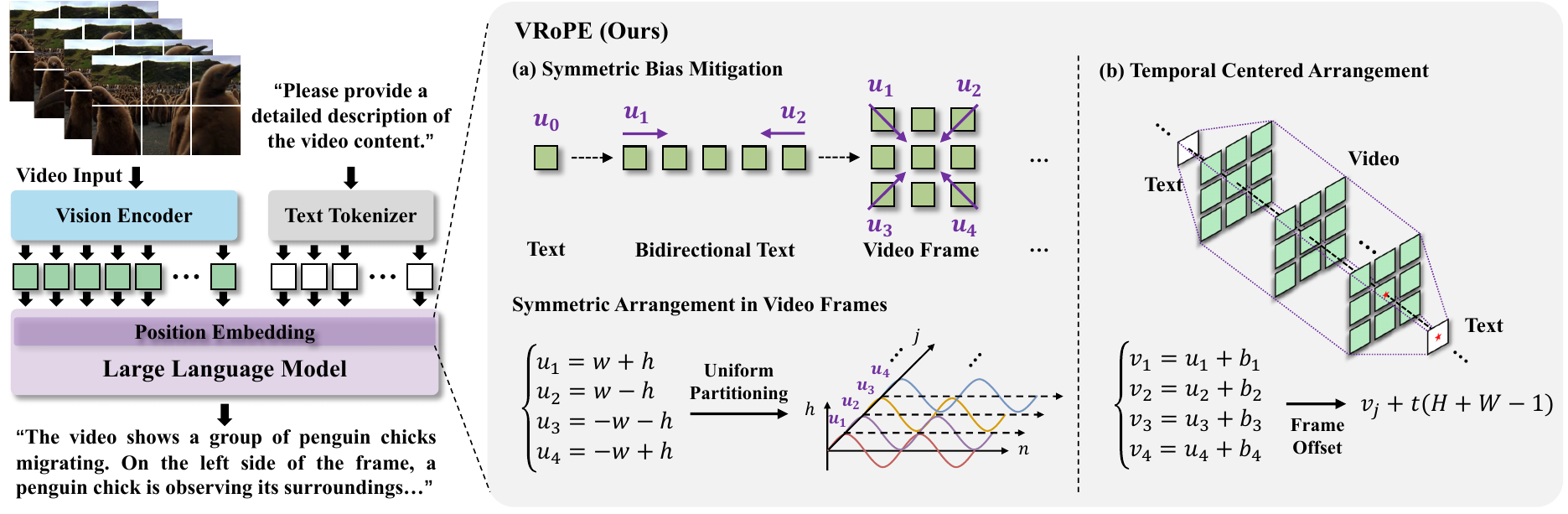}
\caption{\textbf{Left}: the overall architecture of a typical Video-LLM. In this work, our improvements primarily target the positional embedding component of the LLM to enhance its video understanding capability. \textbf{Right}: method illustration of \methodname. \textbf{(a)} We first apply symmetric arrangement to mitigate positional bias in video frames. The RoPE frequencies are uniformly allocated to the four dimensions. \textbf{(b)} We propose to use temporal centered arrangement in video frames to form a seamless video-text transition, which enables video input of arbitrary length without causing discontinuity.}
\label{fig:vrope}
\end{figure*}

\paragraph{(2) Cross-Modal Positional Discontinuity.}
RoPE-3D introduces separate positional encodings for spatial (width, height) and temporal (frame index) dimensions. However, when video tokens are concatenated with subsequent text tokens, their positional indices do not follow a smooth transition. Instead, text tokens inherit positional indices that are arbitrarily offset  by the maximum position value across spatial $(W, H)$ and temporal dimensions $T$, i.e., $\max(W, H, T)$.
This results in an artificial “jump” in the positional encoding space when transitioning from video to text tokens. \textit{The discontinuity creates an abrupt and non-uniform gap between the final video token and the subsequent text token.} As is shown in Table \ref{tab:discontinuity}, text-to-video attention weights of RoPE-3D at the video-text boundary are an order of magnitude lower than RoPE and \methodname, which demonstrates that the discontinuity in position embedding will affect the attention weights. Further, the magnitude of this gap depends on video dimensions rather than being a fixed offset, making it inconsistent across different video-text samples. 
Such a discrepancy can degrade the model’s ability to establish seamless contextual dependencies across modalities. This issue is particularly problematic in long videos, as the increasing frame count $T$ exacerbates the positional gap, which will be further discussed in Section \ref{sec:retrieval}.

\section{Method: \methodname}

In this section, we introduce Video Rotary Position Embedding (\methodname), a novel positional encoding method tailored for Video-LLMs. Our approach addresses the inherent limitations of RoPE-3D, including positional attention bias and cross-modal positional discontinuity, by leveraging a combination of Symmetric Bias Mitigation and Temporal Centered Arrangement.
The overall framework of \methodname is illustrated in Figure \ref{fig:vrope}.

\subsection{Symmetric Bias Mitigation}

As discussed in Section \ref{sec:problem}, both RoPE and RoPE-3D employ a single positional arrangement direction when encoding features within video frames (e.g., row-major scanning for RoPE and top-left to bottom-right ordering for RoPE-3D), inevitably introducing positional attention bias. To address this limitation, we propose Symmetric Bias Mitigation as illustrated in Figure \ref{fig:vrope} (a).

Specifically, we design a unified symmetric positional arrangement paradigm applicable to arbitrary dimensions. For textual tokens represented as points, their inherent symmetry is preserved. For one-dimensional sequences, we adopt bidirectional positional indexing starting from both endpoints (similar to bidirectional modeling in language models). For two-dimensional planes (i.e., video frames), we implement a four-directional symmetric arrangement extending from frame vertices toward the center. This scheme naturally extends to three-dimensional space with eight-vertex symmetry, etc. Given an input video frame of size $(W, H)$, we compute four symmetric directional positional arrangements as follows:

\begin{equation}
\label{eq:symmetric}
\begin{pmatrix}
u_{1} \\
u_{2} \\
u_{3} \\
u_{4}
\end{pmatrix} = 
\begin{pmatrix}
w + h \\
w - h \\
-w - h \\
-w + h
\end{pmatrix}.
\end{equation}

Considering that RoPE employs different frequencies across channels, we strategically allocate frequencies to these four symmetric positional indices in a uniform manner. This design enables distinct positional arrangement directions to model features through different RoPE frequencies (high, medium and low).

\subsection{Temporal Centered Arrangement}

While Symmetric Bias Mitigation effectively alleviates positional bias, the inherent discontinuity between video and textual modalities persists. To address this challenge, we propose the Temporal Centered Arrangement for positioning video frames. Given that textual positions inherently satisfy $u_{1} = u_{2} = u_{3} = u_{4}$ (demonstrating isotropic symmetry), we first align the geometric center of each video frame with the textual arrangement axis through coordinate transformation. Specifically, for a video of size $(W, H, T)$ with an initial position index $p_{start}$ (i.e., the last position id + 1 of the previous text), this process can be denoted as:

\begin{equation}
\label{eq:center}
\begin{pmatrix}
v_{1} \\
v_{2} \\
v_{3} \\
v_{4}
\end{pmatrix} = 
\begin{pmatrix}
u_{1} \\
u_{2} + H - 1 \\
u_{3} + H + W -2 \\
u_{4} + W - 1
\end{pmatrix} + p_{start}.
\end{equation}

Subsequently, we systematically arrange frame positions along the temporal dimension using the following formulation:

\begin{equation}
\label{eq:offset}
v_{j}^{t} = v_{j} + t(H + W - 1),
\end{equation}

\noindent where $t$ is the frame index. This configuration ensures that: (1) The central position of each video frame coincides with the textual arrangement axis, and (2) Sequential frames naturally extend along the textual positional progression direction through temporal ordering. Consequently, the temporal expansion axis of video sequences becomes intrinsically aligned with the positional growth direction of text tokens, which means that arbitrary length of video input does not affect the continuity between video and text.

Finally, our \methodname computes the positional encoding as:
\begin{equation}
\begin{aligned}
\text{VRoPE}_{j}(& \mathbf{x}, v_{1}^{t}, v_{2}^{t}, v_{3}^{t}, v_{4}^{t}) \\
& = \left\{
\begin{aligned}
& \text{RoPE}_{j}(\mathbf{x}, v_{1}^{t})_{j = 4k} \\
& \text{RoPE}_{j}(\mathbf{x}, v_{2}^{t})_{j = 4k + 1} \\
& \text{RoPE}_{j}(\mathbf{x}, v_{3}^{t})_{j = 4k + 2} \\
& \text{RoPE}_{j}(\mathbf{x}, v_{4}^{t})_{j = 4k + 3}
\end{aligned}
\right.
\end{aligned}
\end{equation}
\noindent where $k \in \left\{0, 1, 2, ...\right\}$. For text tokens, we retain the original RoPE encoding structure (Eq. \ref{eq:text_inv}) to ensure compatibility with LLMs. Further discussions can be found in Appendix \ref{sec:discussion}.

\section{Experiments}

\subsection{Experimental Setup}

\paragraph{Implementation Details.} 
We apply our proposed \methodname to Video-LLM architectures with three widely used LLM backbones: Vicuna-7B, Qwen2-1.5B, and Qwen2-7B, the resulting models are denoted as Video-Vicuna-7B, Video-Qwen2-1.5B, and Video-Qwen2-7B. For the vision encoder, we leverage Eva-CLIP \cite{sun2023eva}, and connect the Vision Encoder to the LLM using a Multi-Layer Perceptron (MLP) \cite{tolstikhin2021mlp}. We use a $224 \times 224$ resolution for both image and video inputs. For video input, the number of input frames is 16 and the frames are tokenized using a $2 \times 2$ pooling kernel with a stride of 2, i.e., each frame has 64 tokens as input. Training follows a two-stage paradigm: in the pre-training stage, only the MLP connector is trained, while in the instruction-tuning stage, both the MLP and LLM backbones are fine-tuned, with the Vision Encoder frozen throughout.
During pre-training, we use a batch size of 256 and a learning rate of 1e-3, while for instruction-tuning, we reduce the batch size to 128 and set the learning rate to 2e-5. A warm-up ratio of 0.03 is used, followed by cosine learning rate decay after the linear warm-up phase. The training was conducted on 8 Nvidia A800 GPUs.

\paragraph{Training Data.}
For Vicuna-7B, we pre-train the model on the LLaVA-558K dataset \cite{liu2024visual} with WebVid samples \cite{bain2021frozen} and fine-tune it on the LLaVA-mix665K \cite{liu2024visual} dataset augmented with VideoChatGPT data \cite{maaz2023video}. 
For the Qwen2 LLM series, we pre-train the models on a randomly sampled 1M caption dataset, which includes LLaVA-558K, WebVid, DenseFusion-1M \cite{li2024densefusion}, VALOR \cite{liu2024valor}, and CC3M \cite{changpinyo2021conceptual}. The models are then fine-tuned on a combination of LLaVA-mix665K, VideoChatGPT, and LLaVA-Video-178K \cite{zhang2024video}.

\paragraph{Evaluation Benchmarks.}
We evaluated \methodname across diverse video benchmarks, covering \textit{general video understanding} (Video-MME \cite{fu2024video}), \textit{video temporal understanding} (MVBench \cite{li2024mvbench}, TempCompass \cite{liu2024tempcompass}), \textit{long video understanding} (MLVU \cite{zhou2024mlvu}, LongVideoBench \cite{wu2025longvideobench}, EgoSchema \cite{mangalam2024egoschema}), and \textit{long video retrieval} (Video-NIAH \cite{zhao2024needle}) to validate its effectiveness. The evaluation is conducted using the official code provided by each benchmark.

\begin{table*}[t]
\centering
\caption{Performance comparison of RoPE variants on video benchmarks across different LLM backbones. Results across tasks, including general video understanding (Video-MME), video temporal understanding (MVBench, TempCompass), and long video understanding (MLVU, LongVideoBench, EgoSchema).}
\label{tab:main}
\resizebox{\linewidth}{!}{
\begin{tabular}{lccccccc}
\toprule
\multirow{2}{*}{Method} & Video-MME & MLVU & \multirow{2}{*}{MVBench} & LongVideoBench & TempCompass & EgoSchema & \multirow{2}{*}{Avg.} \\
& (w/o subs) & @M-Avg && @Val & @Multi-Choice & @Test & \\
\midrule
\textbf{Video-Vicuna-7B} \\
\ \ w/ RoPE & 38.5 & 47.00 & 43.90 & 41.66 & 53.10 & 35.92 & 43.35 \\
\ \ w/ RoPE-3D & 38.0 (\decreasingData{0.5}) & 46.30 (\decreasingData{0.7}) & 44.55 (\increasingData{0.65}) & 40.16 (\decreasingData{1.5}) & 54.94 (\increasingData{1.84}) & 39.79 (\increasingData{3.87}) & 43.96 (\increasingData{0.61}) \\
\tableLineColor
\ \ w/ \methodname & 38.9 (\increasingData{0.4}) & 47.37 (\increasingData{0.37}) & 45.18 (\increasingData{1.28}) & 40.69 (\decreasingData{0.97}) & 54.05 (\increasingData{0.95}) & 40.71 (\increasingData{4.79}) & 44.48 (\increasingData{1.13}) \\
\midrule
\textbf{Video-Qwen2-1.5B} \\
\ \ w/ RoPE & 39.0 & 51.15 & 51.15 & 46.63 & 56.96 & 48.50 & 48.90 \\
\ \ w/ RoPE-3D & 39.3 (\increasingData{0.3}) & 51.19 (\increasingData{0.04}) & 50.45 (\decreasingData{0.70}) & 48.01 (\increasingData{1.38}) & 57.97 (\increasingData{1.01}) & 49.00 (\increasingData{0.50}) & 49.32 (\increasingData{0.42}) \\
\tableLineColor
\ \ w/ \methodname & 42.4 (\increasingData{3.4}) & 51.76 (\increasingData{0.61}) & 50.78 (\decreasingData{0.37}) & 47.79 (\increasingData{1.16}) & 57.15 (\increasingData{0.19}) & 49.90 (\increasingData{1.40}) & 49.96 (\increasingData{1.06}) \\
\midrule
\textbf{Video-Qwen2-7B} \\
\ \ w/ RoPE & 50.1 & 54.87 & 54.33 & 49.36 & 63.73 & 57.14 & 54.92 \\
\ \ w/ RoPE-3D & 49.5 (\decreasingData{0.6}) & 56.06 (\increasingData{1.19}) & 54.23 (\decreasingData{0.1}) & 49.55 (\increasingData{0.19}) & 64.49 (\increasingData{0.76}) & 58.80 (\increasingData{1.66}) & 55.44 (\increasingData{0.52}) \\
\tableLineColor
\ \ w/ \methodname & 50.6 (\increasingData{0.5}) & 57.81 (\increasingData{2.94}) & 54.70 (\increasingData{0.37}) & 50.48 (\increasingData{1.12}) & 65.88 (\increasingData{2.15}) & 58.60 (\increasingData{1.46}) & 56.35 (\increasingData{1.43}) \\
\bottomrule
\end{tabular}
}
\end{table*}

\begin{figure*}[t]
\centering
\includegraphics[width=\linewidth]{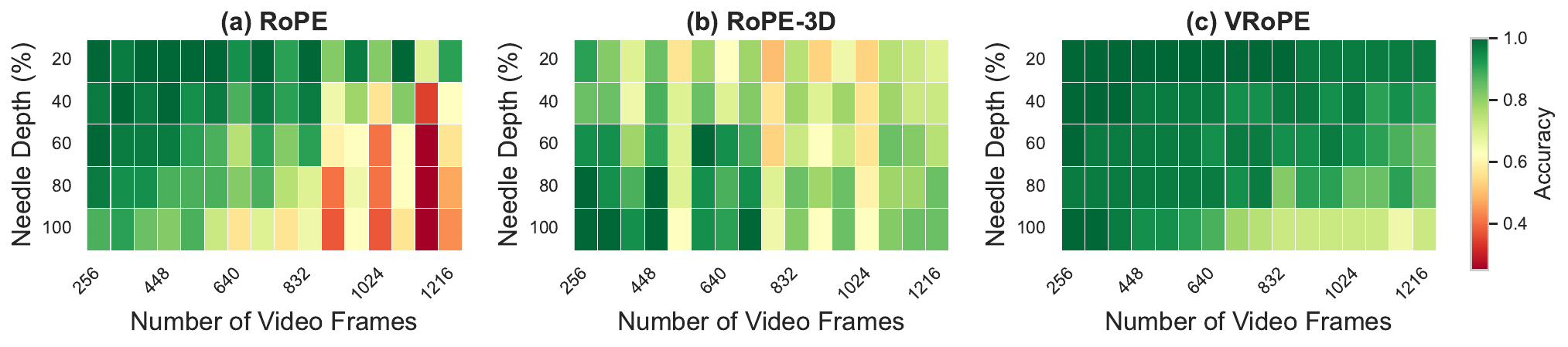}
\caption{Visualization of long video retrieval results on Video-NIAH \cite{zhao2024needle}. Our \methodname consistently achieves high accuracy across varying background lengths and needle depths, showing strong retrieval capability in long videos.}
\label{fig:video_niah}
\end{figure*}

\subsection{Main Results}
\label{sec:main}

We evaluate the performance of RoPE, RoPE-3D, and our proposed \methodname across six video understanding benchmarks. As shown in Table \ref{tab:main}, \methodname consistently outperforms both RoPE and RoPE-3D, achieving the highest average scores across all tasks and backbones.

For instance, in the Video-Vicuna-7B row, \methodname achieves an average score of 44.48, surpassing RoPE by 1.13 points. Similarly, when evaluated with Qwen2-1.5B and Qwen2-7B, \methodname demonstrates consistent improvements across all benchmarks. Notably, it outperforms RoPE and RoPE-3D by significant margins on tasks such as Video-MME (a 3.4-point increase for Qwen2-1.5B) and MLVU (a 2.94-point increase for Qwen2-7B).

These results highlight the superior adaptability of \methodname across different LLM types and parameter sizes. Importantly, \methodname introduces no new learnable parameters and does not increase computational complexity, making it a cost-free performance enhancement for Video-LLMs. More results and visualization examples can be found in Appendix \ref{sec:more_results} and Appendix \ref{sec:visualization}.

\begin{table*}[thbp]
\centering
\caption{We assess various RoPE designs to validate the necessity of the three desired properties: Spatiotemporal Structure Modeling (S.S.M), Positional Unbiasedness (P.U.), and Seamless Video-Text Transition (S.V.T.). The results indicate that the model attains optimal performance when all properties are fully incorporated.}
\label{tab:why_rope}
\resizebox{\linewidth}{!}{
\begin{tabular}{lccccccc}
\toprule
Method & S.S.M. & P.U. & S.V.T. & Video-MME & EgoSchema & LongVideoBench & Avg. \\
\midrule
RoPE & \ding{56} & \ding{56} & \ding{52} & 39.0 & 48.50 & 46.63 & 44.71 \\
RoPE-2D & \ding{52} & \ding{56} & \ding{52} & 43.2 (\increasingData{4.2}) & 47.60 (\decreasingData{0.90}) & 46.33 (\decreasingData{0.30}) & 45.71 (\increasingData{1.00}) \\
RoPE-3D & \ding{52} & \ding{56} & \ding{56} & 39.3 (\increasingData{0.3}) & 49.00 (\increasingData{0.50}) & 48.01 (\increasingData{1.38}) & 45.44 (\increasingData{0.73}) \\
RoPE-Share & \ding{56} & \ding{52} & \ding{52} & 39.7 (\increasingData{0.7}) & 48.66 (\increasingData{0.16}) & 45.10 (\decreasingData{1.53}) & 44.49 (\decreasingData{0.22}) \\
RoPE-Compact & \ding{52} & \ding{56} & \ding{52} & 38.1 (\decreasingData{0.9}) & 50.77 (\increasingData{2.27}) & 45.96 (\decreasingData{0.67}) & 44.94 (\increasingData{0.23}) \\
\tableLineColor
\methodname & \ding{52} & \ding{52} & \ding{52} & 42.4 (\increasingData{3.4}) & 49.90 (\increasingData{1.40}) & 47.79 (\increasingData{1.16}) & 46.70 (\increasingData{1.99}) \\
\bottomrule
\end{tabular}
}
\end{table*}

\begin{table}[t]
\centering
\caption{Average retrieval accuracy across different input frame length intervals on Video-NIAH \cite{zhao2024needle}. Compared to RoPE, the performance advantage of \methodname becomes more pronounced at longer video lengths.}
\label{tab:video_niah}
\resizebox{\linewidth}{!}{
\begin{tabular}{lcccc}
\toprule
Method & 256-512 & 512-768 & 768-1024 & 1024-1216 \\
\midrule
RoPE & 94.84 & 87.03 & 73.28 & 54.84 \\
RoPE-3D & 88.90 & 80.94 & 69.69 & 72.81 \\
\tableLineColor
\methodname & \textbf{98.28} & \textbf{95.16} & \textbf{90.31} & \textbf{87.03} \\
\bottomrule
\end{tabular}
}
\end{table}

\subsection{Results on Long Video Retrieval}
\label{sec:retrieval}
We compare our method with RoPE \cite{su2024roformer} and RoPE-3D \cite{wang2024qwen2} on the long video retrieval task to evaluate the model’s generalization ability with longer video inputs. Following the setup in Video-NIAH \cite{zhao2024needle}, we conduct Video Needle-In-A-Haystack (V-NIAH) experiments, where a target "needle" frame is inserted into a sequence of background frames, with the total frame count varying between 256 and 1216. 

As shown in Figure \ref{fig:video_niah}, the retrieval accuracy of RoPE drops significantly when the number of input frames exceeds 832, while \methodname outperforms other approaches by a considerable margin. The quantitative results, presented in Table \ref{tab:video_niah}, further evidence this finding. Specifically, \methodname achieves an accuracy that is 32.19 points higher than RoPE and 14.22 points higher than RoPE-3D when the number of input frames increases to 1024-1216. Notably, these results are obtained even though the input frame count in this range is dozens of times greater than the maximum number seen during training. This demonstrates the exceptional extrapolation ability of \methodname. Moreover, RoPE-3D underperforms the RoPE baseline for inputs of 256-512, 512-768, and 768-1024 frames, which further proves that the cross-modal positional discontinuity affects the model's ability to understand videos of different lengths.

\subsection{Ablation Studies}
\label{sec:ablation}

\paragraph{Comparison of RoPE Variants.}
We conduct experiments to assess the impact of three key properties: Spatiotemporal Structure Modeling (S.S.M.), Positional Unbiasedness (P.U.), and Seamless Video-Text Transition (S.V.T.), as discussed in Section \ref{sec:intro}. The results, summarized in Table \ref{tab:why_rope}, highlight the importance of these properties.

We first compare RoPE-2D \cite{agrawal2024pixtral} and RoPE-3D \cite{wang2024qwen2} with the baseline RoPE \cite{su2024roformer} method. \textit{RoPE-2D} encodes only the spatial coordinates $(w, h)$ of each frame. While it resolves the cross-modal positional discontinuity, it still suffers from positional bias. Both RoPE-2D and RoPE-3D show improvements over RoPE, demonstrating the benefits of spatiotemporal structure modeling.

Next, we evaluate two additional variants, \textit{RoPE-Share} and \textit{RoPE-Compact}, to further ablate the impact of S.S.M. and P.U. RoPE-Share uses identical positional embeddings within each frame, arranged sequentially. While it resolves positional bias and ensures continuity, it neglects the spatial structure of the frames, leading to a performance drop compared to RoPE. RoPE-Compact is an extention of RoPE-3D that addresses positional discontinuity by encoding subsequent text tokens with $(W + 1, H + 1, T + 1)^{T}$, but it deviates from text compatibility requirements, which slightly limits its performance. 
In contrast, our proposed method (\methodname) incorporates all three properties, achieving a 1.99-point improvement over the RoPE baseline, surpassing all other variants. More detailed illustration of RoPE-Share and RoPE-Compact can be found in Appendix \ref{sec:other_ropes}.

\begin{table}[t]
\centering
\caption{Ablation study on \methodname components. We evaluate the impact of Symmetric Bias Mitigation (Symmetric) and Temporal Centered Arrangement (Continuity). The model achieves the best performance when both components are applied together.}
\label{tab:ablation}
\resizebox{\linewidth}{!}{
\begin{tabular}{cccc}
\toprule
Continuity & Symmetric & Video-MME & LongVideoBench \\
\midrule
\ding{56} & \ding{56} & 39.0 & 46.63 \\
\ding{52} & \ding{56} & 42.3 & 46.30 \\
\ding{56} & \ding{52} & 41.3 & 47.27 \\
\tableLineColor
\ding{52} & \ding{52} & \textbf{42.4} & \textbf{47.79} \\
\bottomrule
\end{tabular}
}
\end{table}

\paragraph{Ablation on \methodname Components.}
We conduct ablation experiments to evaluate the individual contributions of the Symmetric Bias Mitigation and Temporal Centered Arrangement components. The results, presented in Table \ref{tab:ablation}, reveal that when applied separately, each method produces mixed effects. 
Specifically, Temporal Centered Arrangement improves performance on Video-MME, indicating its effectiveness in enhancing smooth translation for general video understanding. Symmetric Bias Mitigation shows a significant gain on LongVideoBench, indicating its effectiveness in reducing bias in long video tasks. 
When combined in \methodname, the two components work synergistically, resulting in more consistent performance. 

\section{Conclusion}
In conclusion, we propose \methodname, a dedicated positional encoding strategy for Video-LLMs that balances spatiotemporal structure, mitigates attention bias, and ensures a smooth transition between video and text tokens. Extensive experiments on different model scales validate its superior performance in video understanding, temporal reasoning, and retrieval tasks. We believe \methodname can serve as a useful building block for future Video-LLMs, enabling better video-language understanding.

\section{Acknowledgments}

This research is supported by Artificial Intelligence-National Science and Technology Major Project (2023ZD0121200) and the National Natural Science Foundation of China (6243000159, 62102416), and the Key Research and Development Program of Jiangsu Province under Grant BE2023016-3, and CCF-Tencent Rhino-Bird Open Research Fund.

\section{Limitations}
While \methodname demonstrates strong performance, there are some limitations. Due to computational resource constraints, our experiments were limited to models with 1.5B, 7B and 8B (shown in Appendix \ref{sec:more_results}) parameters. Larger-scale models could potentially yield further performance gains. Additionally, although \methodname is adaptable across different dimensions, its extension to other modalities (e.g., audio, 3D point clouds, Electroencephalography (EEG)) and higher-dimensional data (e.g., 4D spatiotemporal or medical imaging data) remains an area for future research and validation.

\bibliography{custom}
\newpage
\appendix

\section{Discussion}
\label{sec:discussion}

\paragraph{Dimensional Adaptability.} A key advantage of \methodname is its ability to degenerate into lower-dimensional embeddings without altering its fundamental structure. Unlike methods that assign separate feature channels for each coordinate, \methodname employs linear combinations of the original coordinates, allowing any dimension set to 1 to seamlessly adapt into lower-dimension form. For instance, when $H = 1$, the encoded positions simplify to $(w, w, -w, -w)$, effectively reducing to a 1D form—unlike previous methods that rely on separate encodings, such as $(w, 0)$. This property is particularly beneficial for adapting pre-trained model's positional encodings from images (2D) or videos (3D) to data of varying dimensions without disrupting the original encoding scheme. Consequently, models can transfer more effectively across modalities while preserving consistent positional behavior.

\section{More Results}
\label{sec:more_results}

\subsection{Results on EventBench}

\begin{table}[t]
\centering
\caption{Performance comparison of RoPE variants on event-based EventBench \cite{du2024towards}.}
\label{tab:more_benchmarks}
\begin{tabular}{lc}
\toprule
Method & EventBench \\
\midrule
\textbf{Video-Vicuna-7B} \\
\ \ w/ RoPE & 38.97 \\
\ \ w/ RoPE-3D & 39.33 (\increasingData{0.36}) \\
\tableLineColor
\ \ w/ \methodname & 40.38 (\increasingData{1.41}) \\
\midrule
\textbf{Video-Qwen2-1.5B} \\
\ \ w/ RoPE & 53.31 \\
\ \ w/ RoPE-3D & 52.76 (\decreasingData{0.55}) \\
\tableLineColor
\ \ w/ \methodname & 54.23 (\increasingData{0.92}) \\
\midrule
\textbf{Video-Qwen2-7B} \\
\ \ w/ RoPE & 59.25 \\
\ \ w/ RoPE-3D & 58.61 (\decreasingData{0.64}) \\
\tableLineColor
\ \ w/ \methodname & 60.35 (\increasingData{1.1}) \\
\bottomrule
\end{tabular}
\end{table}

The benchmark evaluated in Section \ref{sec:main} already encompasses comprehensive capabilities required for video understanding tasks. To further validate temporal reasoning performance, we conduct additional evaluations focusing on event-based tasks involving complex temporal dependencies. As shown in Table \ref{tab:more_benchmarks}, our \methodname demonstrates consistent improvements across all models compared to existing methods. These results confirm that our approach maintains superior comprehension capabilities when processing videos containing intricate event sequences.

\subsection{Results on Video-MME with varying lengths}

\begin{table}[t]
\centering
\caption{Detailed performance comparison of RoPE variants on Video-MME \cite{fu2024video}.}
\label{tab:different_len}
\resizebox{\linewidth}{!}{
\begin{tabular}{lccc}
\toprule
Method & Short & Medium & Long \\
\midrule
\textbf{Video-Vicuna-7B} \\
\ \ w/ RoPE & 46.4 & 38.0 & 31.0 \\
\ \ w/ RoPE-3D & 46.0 (\decreasingData{0.4}) & 37.5 (\decreasingData{0.5}) & 30.6 (\decreasingData{0.4}) \\
\tableLineColor
\ \ w/ \methodname & 46.4 (-) & 38.3 (\increasingData{0.3}) & 31.8 (\increasingData{0.8}) \\
\midrule
\textbf{Video-Qwen2-1.5B} \\
\ \ w/ RoPE & 47.4 & 37.6 & 32.2 \\
\ \ w/ RoPE-3D & 47.1 (\decreasingData{0.3}) & 37.0 (\decreasingData{0.6}) & 33.8 (\increasingData{1.6}) \\
\tableLineColor
\ \ w/ \methodname & 50.1 (\increasingData{2.7}) & 39.3 (\increasingData{1.7}) & 37.8 (\increasingData{5.6}) \\
\midrule
\textbf{Video-Qwen2-7B} \\
\ \ w/ RoPE & 60.2 & 47.6 & 42.5 \\
\ \ w/ RoPE-3D & 60.0 (\decreasingData{0.2}) & 46.7 (\decreasingData{0.9}) & 41.7 (\decreasingData{0.8}) \\
\tableLineColor
\ \ w/ \methodname & 60.4 (\increasingData{0.2}) & 47.6 (-) & 43.9 (\increasingData{1.4}) \\
\bottomrule
\end{tabular}
}
\end{table}

In this section, we analyze the performance of RoPE, RoPE-3D, and our \methodname across varying input video lengths on the Video-MME dataset, as summarized in Table \ref{tab:different_len}. The results indicate that \methodname demonstrates marked superiority in processing long-form videos, while also achieving moderate advantages for medium and short videos, maintaining comparable performance to baselines at minimum. This further validates the effectiveness of our approach in enhancing model comprehension capabilities across varying video durations. The consistent improvements underscore our method's robustness in understanding tasks under various video context lengths.

\subsection{Results under Challenging Conditions}

\begin{table}[t]
\centering
\caption{Results on Video-MME \cite{du2024towards} under lower frame rates (8 frames).}
\label{tab:low_frame}
\begin{tabular}{lc}
\toprule
Method & Acc. \\
\midrule
\textbf{Video-Qwen2-1.5B} \\
\ \ w/ RoPE & 38.9 \\
\ \ w/ RoPE-3D & 37.2 (\decreasingData{1.7}) \\
\tableLineColor
\ \ w/ \methodname & 40.9 (\increasingData{2.0}) \\
\bottomrule
\end{tabular}
\end{table}

\begin{figure*}[t]
\centering
\includegraphics[width=0.9\linewidth]{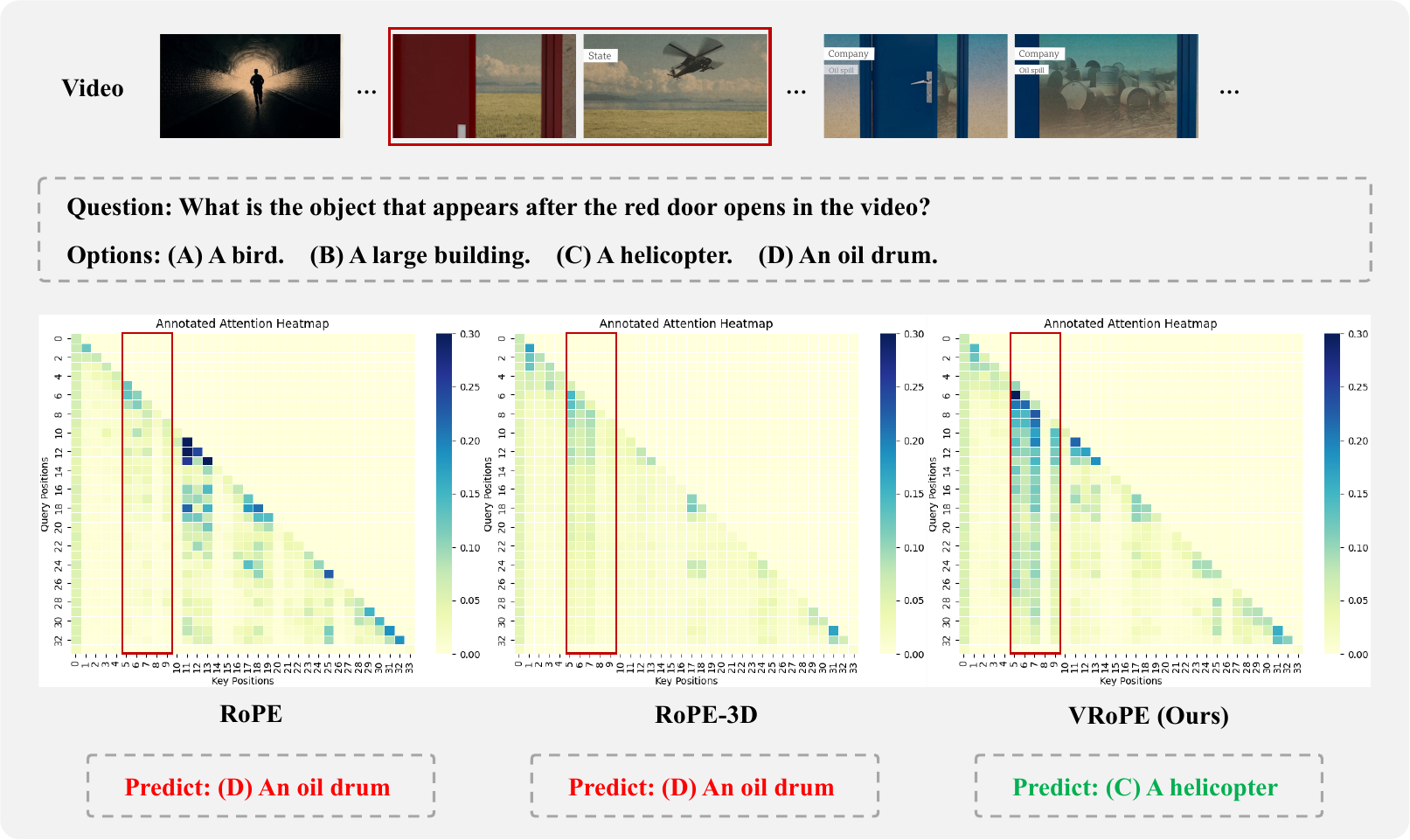}
\caption{Attention weight visualization of RoPE, RoPE-3D, and \methodname. The visualization reveals that \methodname exhibits stronger attention activation within critical frames (highlighted by red boxes), demonstrating its accurate focus on pivotal spatiotemporal regions. In contrast, RoPE and RoPE-3D display attenuated attention responses in these corresponding areas, indicating insufficient awareness of key events. This attention misalignment consequently leads to erroneous predictions, as evidenced by their incorrect interpretations of the visual content.}
\label{fig:visualization}
\end{figure*}

In this section, we evaluate the performance of RoPE, RoPE-3D, and \methodname on Video-MME under low frame-rate inputs (8 frames), as reported in Table \ref{tab:low_frame}. Notably, \methodname maintains enhanced performance even in these challenging sparse-sampling scenarios, empirically confirming the robustness of our approach. This empirical evidence highlights our method's capability to preserve spatiotemporal coherence under severe input degradation.

\begin{table*}[t]
\centering
\caption{Performance comparison of RoPE variants on larger models and datasets. Results across tasks, including general video understanding (Video-MME), video temporal understanding (MVBench, TempCompass), and long video understanding (MLVU, LongVideoBench, EgoSchema).}
\label{tab:scale_up}
\resizebox{\linewidth}{!}{
\begin{tabular}{lccccccc}
\toprule
\multirow{2}{*}{Method} & Video-MME & MLVU & \multirow{2}{*}{MVBench} & LongVideoBench & TempCompass & EgoSchema & \multirow{2}{*}{Avg.} \\
& (w/o subs) & @M-Avg && @Val & @Multi-Choice & @Test & \\
\midrule
\textbf{Video-Qwen3-8B} \\
\ \ w/ RoPE & 61.00 & 64.96 & 59.68 & 60.81 & 68.67 & 56.41 & 61.92 \\
\ \ w/ RoPE-3D & 61.44 (\increasingData{0.44}) & 64.50 (\decreasingData{0.46}) & 59.34 (\decreasingData{0.34}) & 61.00 (\increasingData{0.19}) & 69.11 (\increasingData{0.44}) & 56.03 (\decreasingData{0.38}) & 61.90 (\decreasingData{0.02}) \\
\tableLineColor
\ \ w/ \methodname & 62.56 (\increasingData{1.56}) & 65.36 (\increasingData{0.40}) & 59.23 (\decreasingData{0.45}) & 61.48 (\increasingData{0.67}) & 68.67 (-) & 57.07 (\increasingData{0.66}) & 62.40 (\increasingData{0.48}) \\
\bottomrule
\end{tabular}
}
\end{table*}

\subsection{Results of Larger Models and Datasets}

In this section, we validate the superiority of our approach through scaled-up model architectures and expanded training datasets. Specifically, we conduct experiments using SigLIP-2 \cite{tschannen2025siglip} and Qwen3-8B \cite{yang2025qwen3} as backbone architectures. We expand the number of input frames to 32 and the resolution is set to $384 \times 384$. During the pre-training stage, we utilize LLaVA-558K \cite{liu2024visual} combined with 500K randomly sampled video-text pairs from OpenVid-1M \cite{nan2024openvid}. For instruction tuning, we integrate LLaVA-NeXT-790K \cite{li2024llava}, LLaVA-Video-178K \cite{zhang2024video}, and the full OpenVid-1M dataset. This configuration results in approximately 1 million samples for pre-training and 3 million samples for instruction tuning. As demonstrated in Table \ref{tab:scale_up}, \methodname maintains performance advantages even under these enhanced baseline conditions (larger models, expanded datasets, and stronger baselines). These results further substantiate the generalizability and robustness of our method across diverse architectural scales and data regimes.

\section{Visualization Analysis}
\label{sec:visualization}

In Section \ref{sec:problem}, we analyze the positional attention bias and cross-modal positional discontinuity inherent to RoPE and RoPE-3D. To further substantiate these observations, we provide concrete attention visualization examples in this section. As illustrated in Figure \ref{fig:visualization}, for an input video sequence, our \methodname effectively focuses on the video frames most relevant to the query (the red door and the helicopter), whereas RoPE and RoPE-3D exhibit insufficient attention to critical frames. This deficiency leads to localization errors and subsequent incorrect responses – for instance, misidentifying the opening of a black door as the opening of a red door in this example. The comparative visualization demonstrates our method's enhanced capability in spatiotemporal feature localization and event understanding.

\section{Detailed Illustration of Other RoPE Variants}
\label{sec:other_ropes}

\paragraph{RoPE-Share.}
RoPE-Share is a 1D positional encoding where all spatial tokens within a video frame share the same positional ID, i.e., the positional IDs of all frame tokens in the $t$th frame are $n + t$. Text tokens follow the original encoding: $n + T + 1, n + T + 2, ...$. While this design eliminates spatial attention bias and ensures cross-modal continuity, it fails to model spatial positional relationships within frames, leading to suboptimal performance (as is shown in Section \ref{sec:ablation}).

\paragraph{RoPE-Compact.}
RoPE-Compact is a variant of RoPE-3D. The key difference lies in handling cross-modal boundaries: (1) RoPE-3D assigns the next text token a positional ID of $(\max(W, H, T), \max(W, H, T), \max(W, H, T))^{T}$. For example, if $ T > W, H $, the last video token is $ (W, H, T)^{T} $, and the next text token becomes $ (T, T, T)^{T} $, causing discontinuity in the $ w $ and $ h $ dimensions (as shown in Section \ref{sec:retrieval}). (2) To address the above issue, RoPE-Compact increments each dimension by 1, and uses it as the positional ID for the next text token: $(W + 1, H + 1, T + 1)^{T}$. While this resolves cross-modal discontinuity, it disrupts the pre-trained RoPE’s positional frequency patterns of text, degrading performance.

\section{License Statement}
The scientific artifacts used in this work are all publicly available and this work only uses them for research purposes, thus not violating any of the artifacts' licenses. The new models released in this work is also licensed for research purposes only, prohibiting any other misuse.

\end{document}